\documentclass[letterpaper, 10 pt, conference]{ieeeconf}  

\usepackage{graphics} 
\usepackage{amsmath} 
\usepackage{tikz}
\usetikzlibrary{shapes,arrows}
\usepackage{cite}
\usepackage[ruled,vlined]{algorithm2e}
\usepackage{graphicx}
\graphicspath{ {./images/} } 
\IEEEoverridecommandlockouts                              

\title{\LARGE \bf
OGGN: A Novel Generalized Oracle Guided Generative Architecture for Modelling Inverse Function of Artificial Neural Networks}

\author{Mohammad Aaftab V$^{1}$ and Mansi Sharma$^{2}$

\thanks{$^{1}$Mohammad Aaftab V is a final Year B.Tech. Mechanical Engineering Student at IIT Madras, India.
        {\tt\small aaftaabv@gmail.com}}%
\thanks{$^{2}$Mansi Sharma is with the Department of Electrical Engineering, IIT Madras, India.
{\tt\small mansisharmaiitd@gmail.com}}%
}

\begin{document}

\maketitle
\thispagestyle{empty}
\pagestyle{empty}

\begin{abstract}

This paper presents a novel Generative Neural Network Architecture for modelling the inverse function of an Artificial Neural Network (ANN) either completely or partially. Modelling the complete inverse function of an ANN involves generating the values of all features that corresponds to a desired output. On the other hand, partially modelling the inverse function means generating the values of a subset of features and fixing the remaining feature values. The feature set generation is a critical step for artificial neural networks, useful in several practical applications in engineering and science. The proposed Oracle Guided Generative Neural Network, \textit{dubbed as OGGN}, is flexible to handle a variety of feature generation problems. In general, an ANN is able to predict the target values based on given feature vectors. The OGGN architecture enables to generate feature vectors given the predetermined target values of an ANN. When generated feature vectors are fed to the forward ANN, the target value predicted by ANN will be close to the predetermined target values. Therefore, the OGGN architecture is able to map, inverse function of the function represented by forward ANN. Besides, there is another important contribution of this work. This paper also introduces a new class of functions, defined as constraint functions. The constraint functions enable a neural network to investigate a given local space for a longer period of time. Thus, enabling to find a local optimum of the loss function apart from just being able to find the global optimum. OGGN can also be adapted to solve a system of polynomial equations in many variables. The experiments on synthetic datasets validate the effectiveness of OGGN on various use cases.

\end{abstract}


\section{INTRODUCTION}

Neural networks are known to be great function approximators for a given data, \textit{i.e.}, for a given dataset with features and targets, neural networks are very useful in modelling the function which maps features to targets \cite{hornik1989multilayer,wu2000dynamic}. The inverse function of a dataset, \textit{i.e.}, the function that maps the targets to corresponding features is also very important. Neural networks have been used to solve the inverse problem, especially in design and manufacturing applications \cite{liu2018training,hassan2020artificial,lenaerts2021artificial,xu2021improved,qu2020inverse,sekar2019inverse,so2019simultaneous,kim2018deep}, where the ability to predict features corresponding to a given target value is vital. The modelling of the inverse function can be accomplished by modelling the inverse function of a neural network that maps features to targets. The problem of finding the inverse of a neural network is also crucial in various engineering problems, especially the optimisation ones \cite{may2016neural,hernandez2013inverse,rajesh2015artificial,hattab2014application,krasnopolsky2009neural}. Inverse problems in imaging and computer vision are being solved with the help of deep convolution neural networks \cite{mccann2017review,ongie2020deep,li2020nett,wang2020multi}. Generative Adversarial Networks\cite{goodfellow2014generative} have also been used extensively in solving inverse problems \cite{ren2020learning,dan2020generative,lenninger2017generative,creswell2018inverting,asim2020invertible}.

Our paper introduces a novel neural network architecture called Oracle Guided Generative Neural Networks (OGGN). The proposed OGGN  model the inverse function (\textit{i.e.}, the function from targets to features) using the forward function (\textit{i.e.}, the function from features to targets). We define oracle as the function mapping features to targets. Theoretically, an oracle can either be a mathematical function or a neural network. The generative neural network is responsible for predicting features that correspond to a desired target, \textit{i.e.}, when the predicted features are modelled using oracle, the output must be close to desired target value. The generator's loss function is dependent on the oracle. Generator tries to minimize loss using gradient descent. Thus, the proposed architecture is called Oracle Guided Generative Neural Network.

The primary contribution of the proposed architecture is solving the problem of modelling the inverse of a given function or neural network. OGGN architecture can generate feature vector based on a given oracle and a given target vector value, such that, when the generated feature vector is fed into the oracle, it outputs the target vector close to the predetermined target value. The proposed OGGN architecture can also find feature vectors subject to constraints like fixed range for a few or all of the feature values. In addition, OGGN can also generate feature vectors corresponding to a required target with one or more feature values fixed as constants. It can also be modified to tackle the problem of solving a system of polynomial equations in many variables. All use cases of OGGN are analyzed on synthetic datasets in various experiments.

\begin{figure*}[t]
    \centering
    \includegraphics[width=400pt,height=200pt]{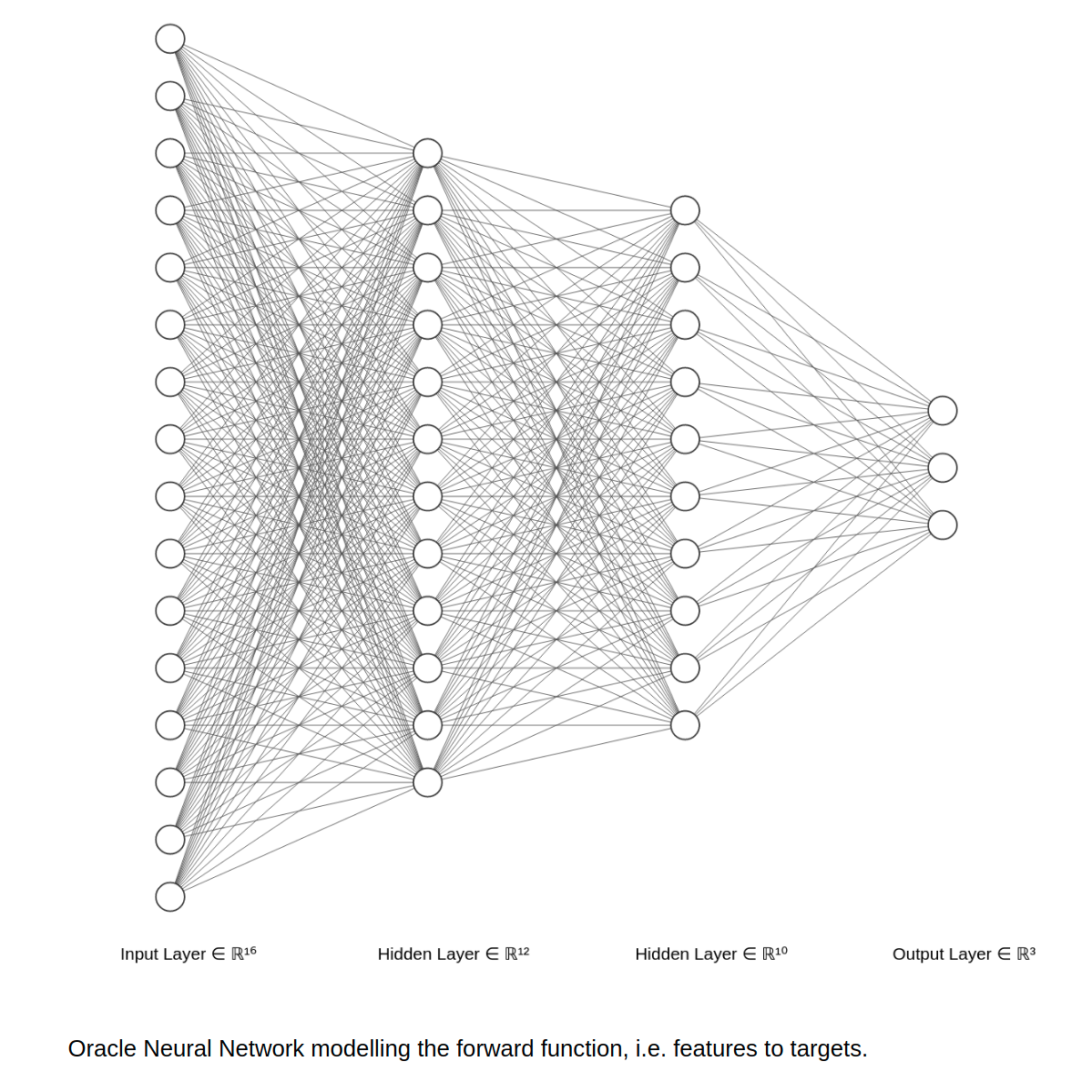}
    \caption{Oracle neural network: maps features to targets}
    \label{fig: props}
\end{figure*}

\begin{figure*}[t]
    \centering
    \includegraphics[width=400pt,height=200pt]{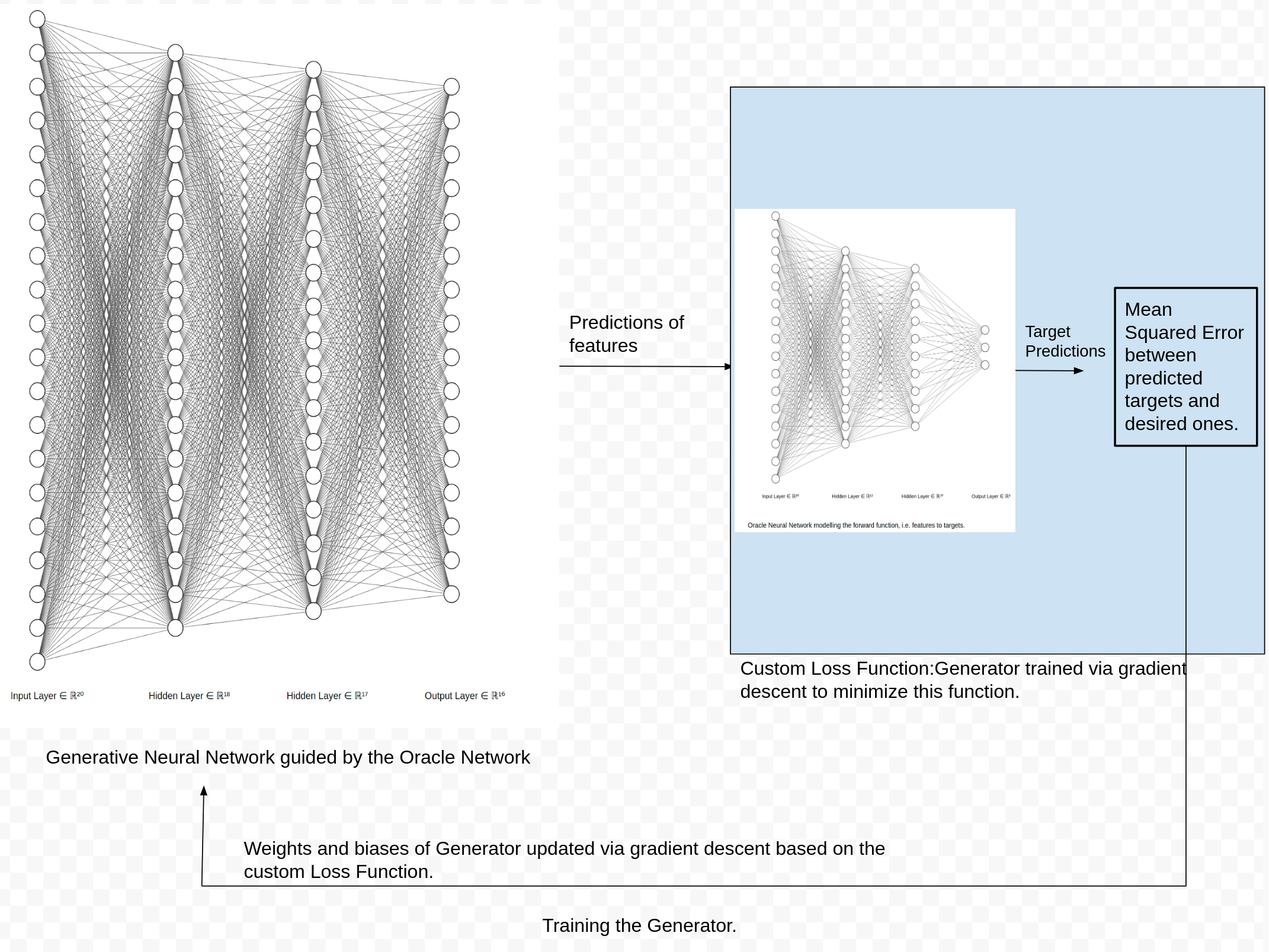}
    \caption{Generative network training methodology}
    \label{fig: genprops}
\end{figure*}

\section{Related Work}

The problem of developing generative neural networks for finding features has been explored well, especially in the recent years. In this section, we explore some latest methods.

Chen et al.\cite{chen2020generative} describes a general‐purpose inverse design approach using generative inverse design networks along with active learning. They introduce a framework named Generative Inverse Design Networks (GIDNs). Their framework has two DNNs namely, the predictor and the generator. The weights and biases of the predictor are learned using gradient decent and backpropagation, while training on a given dataset. In the designer, the weights and biases are adapted from the predictor and set as constants. Initial Gaussian distribution features are fed into the designer as inputs and the optimized design features are generated as outputs. The designer inverse model optimizes the initial inputs or designs based on an objective function via gradient descent with weights and biases of all layers kept as constants and only the input layer values are learned. Active learning takes place in the feedback loop, where the optimized designs or features are verified and added to the dataset. We differ from their approach, in using an additional generator neural network to generate features from random data whose size is tunable according to the complexity of the inverse neural network.

Yang et al.\cite{yang2019adversarial} describe neural network inversion in adversarial settings. They present an approach to generate the training data that were used in training a forward neural network using only black box access to it. Our approach also uses a black box access to the forward neural network, but the use cases, training procedure, and architecture are different. While their approach mainly focuses on generating the exact training data (features and targets) that were used at the time of training the forward neural network, 
our approach shows how to generate features corresponding to any desired target. 

Tahersima et al.\cite{tahersima2019deep} present a method to design integrated photonic power splitters using deep neural networks to model the inverse relation. Their approach deals along the same lines as ours. They also use neural networks to find feature vector that corresponds to a given target value. However, their 
network is different from our approach as it is built for a very specific application of designing integrated photonic power splitters. We proposed an architecture that can be generalized and used for any data. 

Zang et al.\cite{zhang2018multivalued} describes a  multi-valued neural network for inverse modeling and application to microwave filters. They also attempt to find the feature vector corresponding to a required target value. They focus mainly on the existence of non-unique feature vectors corresponding to a given target value. Their approach overcomes the above mentioned non-uniqueness problem by generating multiple sets of feature vectors at a time for a given target vector.

Our proposed methodology uses a separate generator neural network to convert random data into features using oracle neural network. Hence, our approach is different from the previous works.

\section{Proposed Oracle Guided Generative Neural Network}

We introduce a novel neural network architecture known as OGGN, for modelling the inverse of a given neural network. Modelling the inverse of a neural network involves finding the values of features that correspond to a desired target value. The most straight forward way to accomplish this task is to train a new neural network with targets as inputs and features as outputs. The data required for training the new inverse neural network can be obtained by swapping the input and output data used to train the forward network. The straight forward method of modelling the inverse function can often suffer from the multi-valued relationship between inputs and outputs. The multi-valued relationship refers to the fact that different feature vectors can correspond to a same target vector. It means that same input values to the inverse network (targets) can have different outputs (features). This causes significant problems while training the inverse neural network directly from the swapped data. Humayun et al.\cite{kabir2008neural} discusses this problem in detail. 

The proposed OGGN architecture provides a better alternative method for mapping the inverse of a neural network. Oracle network is the name of the forward neural network whose inverse function we are trying to model (figure \ref{fig: props}). The workflow of our architecture is depicted in (figure \ref{fig: genprops}). A generator neural network takes random data as input (the dimensionality of the random data is a hyper-parameter and can be increased or decreased to fit the complexity of the inverse function) and outputs the predicted feature vector value. The training procedure of OGGN is summarized in Algorithm \ref{generalprocd}.

\begin{algorithm}
\label{generalprocd}
\SetAlgoLined
\KwResult{features corresponding to a desired target value}
 random data (size tunable according to the complexity of system of equations)\;
 Oracle Neural Network (maps the features to targets)\;
 \emph{e is the acceptable error between predicted targets and desired targets.}\;
 \While{$loss  > e$}{
  features = generator(random data)\;
  predicted target = oracle(features) \;
  loss = (predicted target$-$desired target$)^{2.0}$\;
  optimizer adjusts weights and biases of generator neural network based on loss\;
  }
 \caption{OGGN Methodology}
\end{algorithm}

Once predicted features from generator are obtained, they are modelled into target vectors using the oracle neural network. The loss value of the generator neural network is the mean squared error between the predicted target vectors and desired target vectors. The training of generator neural network is done via gradient descent to minimize the obtained loss value. Minimizing this loss means that the predicted features are close to the features corresponding to the desired target vector. Thus, as the generator trains, it predicts features closer and closer to the required features (\textit{i.e.}, those that correspond the desired target vector). By training the generator, the values of input features corresponding to a desired value of target vector have been derived. It is possible to generate multiple such input feature vectors at a time by passing many rows into the generator neural network and optimising all of them at the same time. Thus, multiple input feature vectors that all correspond to a same or similar value of a output vector can be found. 

OGGN can solve the following problems:
\begin{itemize}
  \item It is possible to model the inverse function to a given neural network by trying to produce the values of feature vectors that correspond to a required output in the forward neural network.
  \item OGGN can also model only a part of the inverse function of a forward neural network, \textit{i.e.}, it is possible to fix one or more input feature values and train the generator to produce only the remaining feature values such that the whole feature vector corresponds to a desired target value.
  \item OGGN can constraint the values of one or more generated feature values. This is accomplished by using a family of functions introduced in our paper called \emph{constraint functions}.
  \item OGGN can be adapted to solve a system of polynomial equations.
\end{itemize}

All these use cases are validated using synthetic datasets.

\subsection{Constraint Functions}
Activation functions \cite{nwankpa2018activation} in neural networks such as ReLu, Tanh serve various purposes ranging from introducing non-linearity to the modelling process or avoiding gradient vanishing.

We propose one such family of functions named, \emph{constraint functions}, which serve the purpose of constraining the search/exploration space in a neural network. Constraint functions, like activation functions are applied to each output from a neuron. Constraint functions are conditional and they help the neural network minimize the loss while still exploring in a prescribed space. Constraint functions help in maintaining the outputs in a certain range. One such example function is described in the experiments section below.

\section{Experiments and Results}

\subsection{Dataset Preparation}
To test the performance of the proposed model, we generated a synthetic dataset. We consider an arbitrary polynomial exponential function in four variables $x_1, x_2, x_3, x_4$. The function is defined as:
\begin{equation}
y = 9x_1^{0.87}+8.97x_2^{0.02}+0.876x_3^{0.12}+2.9876x_4^{0.987}
\label{polyEq} 
\end{equation}
The training data has been prepared by randomly choosing values of $x_i, i=1,...,4$ variables in (\ref{polyEq}). 
The target variable $y$ is calculated from polynomial function by substituting the randomly selected $x$ values. We consider $10,000$ random sample points of this function (\ref{polyEq}) for training our model. For testing, we generated a new set of $1000$ random samples of the function (\ref{polyEq}).

\subsection{Experimental Settings}

The generated training data is used to train the proposed oracle neural network, which maps the forward function between features and target outputs, \textit{i.e.}, the function mapping from $x$ values to $y$ value. To validate the proposed idea, same oracle network has been used for carrying out all the experiments.
Each experiment is designed to validate a use case of the proposed OGGN architecture. The use cases include: 
\begin{itemize}
  \item Generating values of all features corresponding to a desired target value.
  \item Generating values of some features corresponding to a desired target value, while other feature values are set as constants.
  \item Generating feature values  corresponding to a desired target value, such that, the feature values are restricted within a customizable range.
  \item Solving a system of polynomial equations.
\end{itemize}

Each use case requires a different generator neural network architecture.

\subsection{Use Cases of the Proposed OGGN architecture}

Here, different use cases of proposed OGGN architecture are described and experiments have been performed to validate the proposed model.

\subsubsection{Modelling the inverse function}
The main purpose of this experiment is to generate values of the features corresponding to a desired output. For example, given a target value $y$, our objective is to generate the values of $x_i, i=1,...,4$, such that, when they are substituted in the reference function (\ref{polyEq}), we get the output very near to the previously given target value $y$. In this experiment, we aim to find feature vectors $x_i, i=1,...,4$ corresponding to a required output $y=1900$. So, the generated $x_i, i=1,...,4$ values from this experiment when modelled using the polynomial function (\ref{polyEq}), the output is very close to $1900$. Generalized steps of this use case is summarized in Algorithm \ref{generalprocd}. Pseudo code of this experiment is given in Algorithm \ref{onee}.
\begin{algorithm}
\label{onee}
\SetAlgoLined
\KwResult{features corresponding to a desired target value}
 random data (size tunable according to the complexity of system of equations)\;
 Oracle Neural Network (maps the features to targets)\;
 
 \While{$loss  > e$}{
  $x_1, x_2, x_3, x_4$ = generator(random data)\;
  predicted target = oracle($x_1, x_2, x_3, x_4$) \;
  loss = (predicted target$-$desired target$)^{2.0}$\;
  optimizer adjusts weights and biases of generator neural network based on loss\;
  }
 \caption{Model inverse function using OGGN}
\end{algorithm}
It is not guaranteed to get the exact desired value always, while modelling the generated features using the forward neural network. It may be either because of the error in modelling the polynomial function by the Oracle network, or due to the polynomial function itself never reach the desired value $y$ for any value of $x_i,i=1,...,4$. Hence, the generator network only tries to minimize the error between the desired output and the predicted target. 

After training the generator network for $2000$ epochs, we get the feature values $x_1=224.6277$, $x_2=0.000$, $x_3=283.2135$, $x_4=328.2939$. The target value corresponding to the predicted features is $1911.4099$. It is clear that this predicted target, $1911.4099$ is very close to the desired target value of $1900.00$. With this experiment, we verify that we can compute feature vectors that correspond to an output of our choice. Thus, this experiment demonstrates the feasibility of the proposed model and its principle. It verifies the proof of concept or theoretical underpinnings that the proposed OGGN has potential to solve a number of inverse problems. The feature vectors can be further optimized so that the target can be achieved closer to $1900$. We can explore different approaches like using more data for training of the oracle network or using extended architecture with more layers for oracle and generator neural networks. Training the networks for a larger number of epochs could also improve the accuracy. 

\subsubsection{Finding the other feature values with one feature specified as constant}

In this experiment, we aim to find the optimal values of feature vectors $x_i, i \neq j$, such that one or more variables $x_j$ can be specified as a constant. For example, the value of $x_4$ is set to $10.0$ in the polynomial function (\ref{polyEq}). The objective is to compute the values of $x_1$, $x_2$, $x_3$ using proposed model such that the target output should be close to the desired value $y=2000.0$. The generator predicts the values of $x_1,x_2,x_3$. Now, the feature values (\textit{i.e.}, predicted $x_1, x_2, x_3$ and constant $x_4$) are modelled using oracle network. The oracle gives a predicted output and this prediction is compared to target output via mean squared error. The generator network tries to minimize this error by adjusting its weights and biases using back propagation and gradient descent. After training the generator network for $1000$ epochs, we estimate the feature values $x_1=485.6413$, $x_2=70.6677$, $x_3=90.4546$, $x_4=10.00$. The target value corresponding to the predicted features is $1996.1234$. It is clear that the predicted target value $1996.1234$ is very close to the desire target value of $y=2000.0$. Results can be further improved by using better optimizer or training the networks for longer times with larger datasets.
The pseudo code for training OGGN for this experiment is summarized in Algorithm \ref{twoo}.
\begin{algorithm}
\label{twoo}
\SetAlgoLined
\KwResult{some features corresponding to a desired target value while other features are fixed}
 random data (size tunable according to the complexity of system of equations)\;
 Oracle Neural Network (maps the features to targets)\;
 x4=$10.00$ (this is constant)\;
 
 \While{$loss  > e$}{
  $x_1$, $x_2$, $x_3$ = generator(random data)\;
  predicted target = oracle($x_1$, $x_2$, $x_3$, $x_4$) \;
  loss = (predicted target$-$desired target$)^{2.0}$\;
  optimizer adjusts weights and biases of generator neural network based on loss\;
  }
 \caption{Model subset of inverse function using OGGN}
\end{algorithm}

This experiment has been designed to demonstrate that the proposed OGGN architecture can explore different projections of the inverse function as well. 

\subsubsection{Finding the feature vector with some features in a desired range and some features specified to be constant}

The objective of this experiment is to find the optimal values of feature vectors $x_i, i \neq j$, such that they are in a desired range and one or more variables $x_j$ can be specified as a constant. For instance, the value of $x_4$ is set to $100.0$ in the polynomial function (\ref{polyEq}). The objective is to compute the values of $x_1$, $x_2$, $x_3$ using proposed model, such that, the target output should be close to the desired value $y=788.0$ and the computed values of $x_1$, $x_2$, $x_3$ are within the desired range of $1$ to $100$. The generator predicts the values of $x_1, x_2, x_3$. The training methodology used in this experiment is summarized in Algorithm \ref{three}.

\begin{algorithm}
\label{three}
\SetAlgoLined
\KwResult{some features constrained to a given space corresponding to a desired target value while other features are fixed}
 random data (size tunable according to the complexity of system of equations)\;
 Oracle neural network (maps the features to targets)\;
 $x_4$=$100.00$ (this is constant)\;
 constraints on range of features is given\;
 \emph{constrainedgenerator is the neural network with constraint functions applied to its output layer}\;
 \While{$loss  > e$}{
  $x_1$, $x_2$, $x_3$ = constrainedgenerator(random data)\;
  predicted target = oracle($x_1$, $x_2$, $x_3$, $x_4$) \;
  loss = (predicted target$-$desired target$)^{2.0}$\;
  optimizer adjusts weights and biases of generator neural network based on loss\;
  }
 \caption{Model constrained subset of inverse function using OGGN}
\end{algorithm}

We introduce a family of functions dubbed as constraint functions. These constraint functions decide the range of the output of a neural network. Thus, the feature vectors predicted by generator network are within the predetermined range specified by the constraint functions. The constraint functions are defined as,
\begin{equation}
\left\{\begin{matrix}
x/{c_1} & Upper Bound\leq x \\
x*{c_2} & x\leq Lower Bound 
\end{matrix}\right.
\end{equation}
where, $c_1$ and $c_2$ are constants. The values of $c_1$ and $c_2$ are decided after checking the range of outputs from the generator network, trained without using constraint functions. If the outputs of unconstrained generator network are higher than the required upper bound, it is necessary to divide those outputs with an appropriate number in order to bring outputs into the desired range. This appropriate value is set as constant $c_1$. Similarly, If the outputs of unconstrained generator network are lower than the required lower bound, it is necessary to multiply those outputs with an appropriate number in order to bring outputs into the desired range. This suitable value is set as constant $c_2$.

For example, in this experiment, the generator before using constraint functions produce the outputs in the range of $200-500$. To let the feature values lie between $1$ and $100$, initially $c_1$ is set to $2$. To further increase the exploration time of generator in the range of $1$ and $100$, experimentally, the value of $c_1$ is set to $20.0$. Similarly, the value of $c_2$ has been chosen as $10.0$. Thus, the constraint function is defined as 
\begin{equation}
\left\{\begin{matrix}
x/{20} & 100\leq x \\
x*{10} & x\leq 1 
\end{matrix}\right.
\end{equation}

Like an activation function, the constraint function can be applied to neurons in any layer. However in this experiment, constraint function has been applied on neurons in the output layer only. The constrained generator network predicts $x_1$, $x_2$, $x_3$ in the required range. The value of $x_4$ is set as constant, $x_4=100$. The oracle network models the generated features $x_1$, $x_2$, $x_3$ and $x_4$ and predicts the corresponding target. Final loss is calculated as a Mean Squared Error between the predicted target and the required target. The generator neural network then tries to minimize the final loss via gradient descent. 

The constraint function enables a neural network to search in a prescribed space for a longer period of time. It helps neural network explore the prescribed space thoroughly. If the global minimum of the loss function lies outside the prescribed range, even a constrained neural network will eventually optimize itself to produce outputs that correspond to global minima of loss function. We can arrive at the local optima by stopping training of the constrained neural network once the outputs go beyond the prescribed range.

We obtained output $y=777.5806$ with the predicted feature vector $x_1=100.0423$, $x_2=0.0000$, $x_3=53.7249$, $x_4=100.00$. Here the values of $x_1$, $x_2$, $x_3$ were generated by the generator after training for $200$ epochs. The theoretical maximum value achievable for the function (\ref{polyEq}) is $787.34$, if all the features $x_1$, $x_2$, $x_3$ and $x_4$ are constrained to lie within the range of $1$ and $100$. It is apparent that the predicted value is pretty close to the theoretical maximum value. This experiment is designed to show that proposed OGGN architecture can explore different projections of the inverse function in a subspace. Thus, proposed OGGN architecture can impose constraints on one or more features and find the local optimum of a function.

\subsubsection{Solving a simultaneous system of polynomial equations}
In this experiment, we describe another use case of proposed OGGN architecture. OGGN can be used to solve a system of polynomial equations. It is highly versatile and adaptable. Solving a system of polynomial equations involve finding the values of the variables that satisfy all equations. We consider an arbitrary system of polynomial equations to demonstrate the usefulness of the proposed OGGN architecture.

\begin{equation}
\begin{matrix}
9x^{2}+8.97y^{7.8}+0.876z-32.0=0,
\\ 
12x^{3}+9.97y^{8}+10.876z^{3}-43.0=0
\end{matrix}
\label{SofEq}
\end{equation}
To solve (\ref{SofEq}), we consider the following functions
\begin{equation}
\begin{matrix}
f(x,y,z)=9x^{2}+8.97y^{7.8}+0.876z-32.0,
\\ 
g(x,y,z)=12x^{3}+9.97y^{8}+10.876z^{3}-43.0
\end{matrix}
\label{functionsSofEq}
\end{equation}
Solving the simultaneous system of equations involve finding the values of $x$, $y$, $z$ such that both $f(x,y,z)$ and $g(x,y,z)$ would come close to zero. To this end, we consider the oracle function for generator neural network, given as.

\begin{equation}
    oracle output = (f(x,y,z)^{2}+g(x,y,z)^{2})^{0.5}
    \label{gennetloss}
\end{equation}

The main purpose of oracle is to convert the features into targets. In most real life datasets, a mathematical equation is not available to model the features into targets. The only way to model features into targets in real life datasets is to use a neural network. In the previous experiments, a neural network was used as oracle to demonstrate the ability of the OGGN to work with practical datasets. However, in this example, the objective is to solve a system of equations. A mathematical equation converting variables into targets is readily available. Hence, here the oracle is not a neural network, rather it is a function (\ref{gennetloss}).

The value of the oracle function (\ref{gennetloss}) will be close to zero if and only if both functions $ f(x,y,z)$ and $g(x,y,z)$ are close to zero. Hence the solutions of the system of equations (\ref{SofEq}) are the values of $x$, $y$, $z$ such that upon modelling them using the oracle function (\ref{gennetloss}), the output is close to zero. To this end, we task the generator neural network to find the values of $x$, $y$, $z$ such that the oracle function value (\ref{gennetloss}) will be close to zero. The training process is summarized in Algorithm \ref{solvsys}.
\begin{algorithm}
\label{solvsys}
\SetAlgoLined
\KwResult{solutions of system of equations, $x$, $y$, $z$ }
 random data (size tunable according to the complexity of system of equations)\;
 \While{$loss  > e$}{
  $x$, $y$, $z$ = generator(random data)\;
  predicted target = oracle($x$, $y$, $z$) (\ref{gennetloss})\;
  loss = (predicted target$-0.00)^{2.0}$\;
  optimizer adjusts weights and biases of generator neural network based on loss\;
  }
 \caption{Solving System of Equations (\ref{SofEq}): the OGGN way}
\end{algorithm}

The error e, in algorithm \ref{solvsys} is a small number close to zero. The variables $x$, $y$, $z$ predicted by the generator are modelled using the oracle (\ref{gennetloss}). The corresponding loss is calculated as a mean square error between predicted oracle function value and zero. Generator neural network now tries to minimize the loss via gradient descent and back propagation.

After training generator neural network, it predicts values of $x,y,z$ such that when modelled using oracle (\ref{gennetloss}), we get an output of $0.0589$. The values generated are $x=1.2279$, $y=1.0952$, $z=0.2624$. For these variables, $f(x,y,z)=0.03179$ and $g(x,y,z)=0.04953$.

If the given system of equations has a solution, we can be certain that the generated values $x=1.2279$, $y=1.0952$, $z=0.2624$ are close to the actual solution. If the system does not have an exact solution, we can conclude that the obtained values of variables $x$, $y$, $z$ take both $f(x,y,z)$ and $g(x,y,z)$ as close to zero as possible. Hence, we conclude in this example that OGGN is capable of solving a system of polynomial equations.

\section{Conclusions and Future work}

This paper introduces a novel Oracle Guided Generative Neural Networks (OGGN) to solve the problem of finding the inverse of a given function either fully or partially. The problem of predicting the features corresponding to a desired target is very important and useful in several design and manufacturing fields \cite{liu2018training,hassan2020artificial,lenaerts2021artificial,xu2021improved,qu2020inverse,sekar2019inverse,so2019simultaneous,kim2018deep}. It is also very useful in various engineering problems, especially optimization ones \cite{may2016neural,hernandez2013inverse,rajesh2015artificial,hattab2014application,krasnopolsky2009neural}. OGGN can also generate features subject to constraints, \textit{i.e.}, it can generate features that lie within a given range. Our paper also introduces a class of functions known as \emph{Constraint Functions}. Constraint functions enable a given neural network to explore a given space thoroughly and arrive at a local optima. The proposed concept of OGGN architecture coupled with constraint functions have huge potential in several practical applications in design and manufacturing industries. OGGNs are flexible and can be adapted to solve a large variety of research problems.

This paper has described a way to generate feature vectors that are able to model a particular fixed target value using OGGN. In the future, we can further extend the proposed idea for generating feature vectors corresponding to different target values. This aids in the process of extending a given dataset by creating new synthetic data. The synthetic data can build a better oracle neural network. This improved oracle can guide the generator more effectively and can lead to the creation of better synthetic data. Thus, OGGNs can potentially help in active learning of both forward mapping and inverse mapping neural networks for a given dataset. Synthetic dataset generation using the described architecture could helpful in classification tasks \cite{fawaz2019deep}. Especially, if there is a lot of class imbalance \cite{japkowicz2002class,liu2008exploratory,johnson2019survey}.  OGGNs can create synthetic data corresponding to the lesser represented class and potentially overcome the problem of class data imbalance.

\addtolength{\textheight}{-12cm}   


\bibliographystyle{ieeeconf}
\footnotesize
\bibliography{root}

\end{document}